\documentclass{article}

\usepackage{PRIMEarxiv}

\usepackage[utf8]{inputenc} 
\usepackage[T1]{fontenc}    
\usepackage{hyperref}       
\usepackage{url}            
\usepackage{booktabs}       
\usepackage{amsfonts,xcolor}       
\usepackage{nicefrac}       
\usepackage{microtype}      
\usepackage{lipsum}
\usepackage{fancyhdr}       
\usepackage{graphicx, subcaption}       
\graphicspath{{media/}}     

\title{Interpretable Aneurysm Classification via 3D Concept Bottleneck Models: Integrating Morphological and Hemodynamic Clinical Features
}

\author{
  Toqa Khaled \\
  Biomedical Sciences Program, School of Sciences, Zewail City of Science and Technology  \\
  Multimedia Interaction and Communication Lab, Arab Academy for Science and Technology \\
  \texttt{s-toqa.ashmawy@zewailcity.edu.eg} \\
   \And
  Ahmad Al-Kabbany \\
  Multimedia Interaction and Communication Lab \\
  Wearables, Biosensing, and Biosignal Processing Research lab \\
  Arab Academy for Science and Technology \\
  \texttt{alkabbany@ieee.org, alkabbany@aast.edu} \\
}

\begin{document}
\maketitle

\begin{abstract}
We are concerned with the challenge of reliably classifying and assessing intracranial aneurysms using deep learning without compromising clinical transparency. While traditional black-box models achieve high predictive accuracy, their lack of inherent interpretability remains a significant barrier to clinical adoption and regulatory approval. Explainability is paramount in medical modeling to ensure that AI-driven diagnoses align with established neurosurgical principles. Unlike traditional eXplainable AI (XAI) methods—such as saliency maps, which often provide post-hoc, non-causal visual correlations—Concept Bottleneck Models (CBMs) offer a robust alternative by constraining the model’s internal logic to human-understandable clinical indices.

In this article, we propose an end-to-end 3D Concept Bottleneck framework that maps high-dimensional neuroimaging features to a discrete set of morphological and hemodynamic concepts for aneurysm identification. We implemented this pipeline using a pre-trained 3D ResNet-34 backbone and a 3D DenseNet-121 to extract features from CTA volumes, which were subsequently processed through a soft bottleneck layer representing human-interpretable clinical concepts. The model was optimized using a joint-loss function to balance diagnostic focal loss and concept mean squared error (MSE), validated via stratified five-fold cross-validation.

Our results demonstrate a peak task classification accuracy of 93.33\% $\pm$ 4.5\% for the ResNet-34 architecture and 91.43\% $\pm$ 5.8\% for the DenseNet-121 model. Furthermore, the implementation of 8-pass Test-Time Augmentation (TTA) yielded a robust mean accuracy of 88.31\%, ensuring diagnostic stability during inference. By maintaining an accuracy-generalization gap of less than 0.04, this framework proves that high predictive performance can be achieved without sacrificing interpretability. This work represents significant potential for enhancing clinician trust by providing verifiable "reasoning" paths, ultimately facilitating more transparent and interactive AI-assisted surgical planning in neurovascular care.

\end{abstract}


\section{Introduction}
Intracranial aneurysms (IAs) represent a critical neurovascular pathology characterized by the focal weakening and subsequent dilatation of cerebral arterial walls, most commonly occurring within the circle of Willis. Affecting approximately 2\% to 5\% of the adult population, these lesions often remain clinically silent and asymptomatic until the point of rupture \cite{vlak2011prevalence}. The hemodynamic and structural failure of an IA typically results in subarachnoid hemorrhage (SAH), a catastrophic event associated with a 40–50\% mortality rate and significant long-term neurological disability among survivors \cite{etminan2016unruptured}. Given that the prevalence of detected aneurysms increases markedly with age and the presence of modifiable risk factors such as hypertension and smoking, the ability to reliably identify and classify these anomalies via non-invasive imaging like Computed Tomography Angiography (CTA) is paramount \cite{thompson2015guidelines}. Early and accurate classification not only facilitates timely surgical intervention—which significantly improves clinical outcomes compared to post-rupture treatment—but also serves as the cornerstone for personalized stroke prevention and longitudinal patient management \cite{wiebers2003unruptured}.

The rapid evolution of deep learning and computer vision has introduced transformative possibilities for the automated detection, segmentation, and classification of intracranial aneurysms \cite{you2025diagnosis}. Traditional radiological assessment is increasingly augmented by high-capacity convolutional neural networks (CNNs) and transformer-based architectures that can process high-dimensional volumetric data with superior consistency compared to manual observation \cite{fu2023deep,shu2022machine}. This technological surge is primarily powered by the emergence of large-scale, multimodal datasets—such as the Clinical, Morphological, and Hemodynamic Aneurysm (CMHA) dataset—which integrate traditional 3D CTA scans with biomechanical indices derived from Computational Fluid Dynamics (CFD). By leveraging features such as Wall Shear Stress (WSS) and complex geometric ratios, these AI models can move beyond simple binary identification to provide nuanced risk stratification and morphological characterization. Consequently, the integration of automated pipelines into the neurosurgical workflow represents a critical step toward reducing the diagnostic burden on clinicians and minimizing the incidence of missed or misclassified vascular pathologies.

Despite the impressive predictive accuracy of current deep learning models, their inherent "black-box" nature remains a significant barrier to their integration into neurosurgical practice \cite{kelly2019key}. In high-stakes medical environments, a diagnostic output without a transparent justification is often insufficient for clinical adoption or regulatory approval. Explainability is not merely an auxiliary feature; it is a clinical necessity that ensures AI-driven diagnoses align with established neurovascular principles \cite{amann2020explainability,holzinger2019causability}. For the management of intracranial aneurysms, a surgeon must understand the specific morphological or hemodynamic drivers—such as an unfavorable aspect ratio or critical wall shear stress—that lead a model to flag a lesion as high-risk. Without this transparency, clinicians cannot effectively validate the model’s reasoning against their own expertise, potentially leading to skepticism or, conversely, over-reliance on a system that may be highlighting non-causal visual correlations \cite{tonekaboni2019clinicians}.

Traditional eXplainable AI (XAI) methods, such as saliency maps or Class Activation Mapping (CAM), primarily offer post-hoc visual correlations that highlight which image regions influenced a decision. However, these heatmaps often lack causal depth and can be inconsistent or misleading in clinical settings where precise anatomical reasoning is required. Concept Bottleneck Models (CBMs) represent a paradigm shift by embedding interpretability directly into the architectural design \cite{koh2020concept}. Instead of mapping raw inputs directly to a final label, CBMs first predict a set of human-specified, interpretable clinical concepts \cite{yuksekgonul2022post}. These intermediate representations allow clinicians to inspect, and even intervene upon, the internal logic of the model, ensuring that predictions are anchored to "gold-standard" clinical indices rather than opaque mathematical artifacts. The success of CBM-based frameworks has already been demonstrated across diverse medical domains, including the grading of hepatocellular carcinoma, white blood cell classification, and pediatric appendicitis diagnosis \cite{yoh2023conceptual}. By adopting this approach for intracranial aneurysms, the diagnostic process can be aligned with established neurosurgical parameters—such as vessel geometry and hemodynamics—thereby fostering the trust necessary for AI-assisted surgical planning.

\begin{figure}[t]
    \centering
    \includegraphics[width=\textwidth]{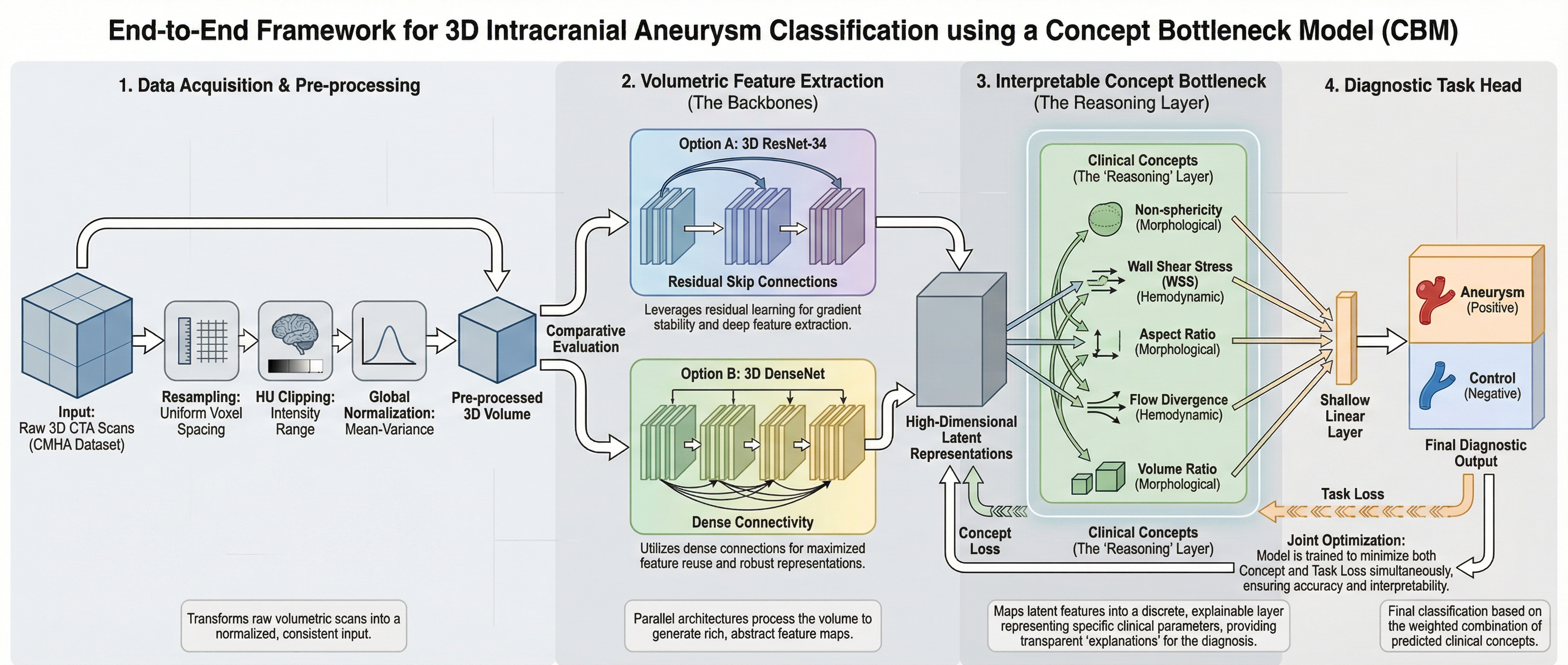}
    \caption{Schematic of the 3D Soft Concept Bottleneck pipeline. Volumetric CTA data are processed through a 3D ResNet-34 or DenseNet-121 backbone to extract latent embeddings. A dedicated head predicts clinical concepts, which are concatenated with visual features to provide an interpretable, multimodal diagnostic output.}
    \label{fig:framework_pipeline}
\end{figure}

In this study, we propose a novel end-to-end 3D Soft Concept Bottleneck Model (CBM) framework specifically tailored for the classification and morphological assessment of intracranial aneurysms. Our architecture leverages high-performance 3D backbones—specifically a pre-trained 3D ResNet-34 and a 3D DenseNet-121 trained from scratch—to extract rich volumetric features from CTA imaging. These latent representations are then mapped to a discrete, interpretable reasoning layer comprising 26 clinical concepts derived from the CMHA dataset. By utilizing a "soft" bottleneck approach, the model concatenates these predicted clinical indices with visual embeddings, ensuring that the final diagnostic output is anchored to established neurosurgical principles without sacrificing the representative power of deep feature extraction. The proposed framework is depcted in in Fig.~\ref{fig:framework_pipeline}, and the primary contributions of this research are summarized as follows:
\begin{enumerate}
    \item Development of a 3D Soft-CBM Architecture: We introduce the first interpretable bottleneck framework for intracranial aneurysms that integrates both 3D morphological and CFD-derived hemodynamic features into a unified diagnostic pathway.

    \item Staged Fine-Tuning and Leakage Mitigation: We implement a rigorous training protocol involving an "unfreezing" strategy for pre-trained encoders and a simple feature selection process to ensure that predicted concepts remain clinically relevant and free from diagnostic data leakage.

    \item Multi-Level 3D Augmentation Strategy: To overcome the constraints of limited medical sample sizes, we propose a three-tier augmentation pipeline—encompassing standard training transformations, strong regularization for synthetic controls, and 8-pass Test-Time Augmentation (TTA)—to maximize model generalization and inference stability.

    \item High-Performance Interpretable Benchmarking: We demonstrate that our framework achieves peak classification accuracies of $93.33\% \pm 4.5\%$ and $91.43\% \pm 5.8\%$ for ResNet and DenseNet backbones, respectively, proving that clinical transparency can coexist with state-of-the-art predictive performance.
\end{enumerate}

The remainder of this article is organized as follows: Section 2 provides a review of related work concerning intracranial aneurysm classification and the evolution of explainable architectures in medical imaging. Section 3 details the methodology, including the characteristics of the CMHA dataset , the volumetric preprocessing pipeline , and the engineering of the Soft Concept Bottleneck architecture. This section also describes the optimization strategies , joint-loss formulations , and the multi-level augmentation protocols employed. Section 4 presents the experimental results, offering a quantitative benchmarking of the 3D backbones , an analysis of training trajectories , and a comparative evaluation of standard versus Test-Time Augmentation (TTA) inference. Finally, Section 5 concludes the paper by summarizing the key findings and discussing potential avenues for future research in interpretable neurovascular care.

\section{Related Work}
\subsection{Recent XAI Methods for Brain-Related Conditions}
The shift toward interpretable neuroimaging has recently seen a transition from static heatmaps to more dynamic, evolutionary, and uncertainty-aware frameworks. The research proposed in \cite{shojaei2023evolutionary} introduced an evolutionary explainable deep learning approach for Alzheimer's MRI classification, which optimizes feature selection to improve both accuracy and transparency in 3D brain scans. This work highlights the importance of identifying specific anatomical biomarkers rather than relying on global visual patterns. Parallelly, the authors of \cite{inamdar2026trustnet} developed TrustNet, a lightweight architecture that integrates uncertainty quantification with quantitative XAI for ischemic stroke detection in CT images. TrustNet is particularly relevant as it addresses the "trust gap" in clinical settings by providing a reliability score alongside visual explanations, a principle that mirrors our use of Test-Time Augmentation (TTA) to ensure diagnostic stability.

\subsection{Recent XAI Methods for Intracranial Aneurysms}
In the neurovascular domain, recent research has focused on the automated detection and risk stratification of lesions. The framework proposed in \cite{you2025diagnosis} presented a comprehensive benchmark for the diagnosis of intracranial aneurysms using deep learning-based detection and segmentation on CTA volumes. While their model achieves high sensitivity, it primarily functions as a "black-box" detection system, emphasizing the need for the clinical bottleneck we propose. Furthermore, the authors of \cite{ou2022morphology} proposed a morphology-aware multi-source fusion-based prediction model for aneurysm rupture. Their approach integrates geometric features with neural embeddings; however, our work extends this concept by using a Soft-CBM architecture that explicitly predicts and concatenates these morphological indices (e.g., aspect ratio, vessel angles) to provide a verifiable reasoning path rather than a fused latent representation.

\subsection{Recent CBMs for Brain-Related Conditions}
Concept Bottleneck Models (CBMs) represent the current frontier in "ante-hoc" interpretability. The study presented in \cite{yoh2023conceptual} demonstrated the clinical utility of this approach by developing a conceptual classification of resectability for hepatocellular carcinoma. Although their work focuses on oncological surgery, it establishes a vital precedent for using human-interpretable clinical concepts to gate a model's final decision. By anchoring the diagnostic output to a "gold-standard" conceptual layer, they proved that such models can match the performance of standard CNNs while providing a transparent audit trail. Our research builds upon this foundation by applying the CBM paradigm to 3D neurovascular volumes, curating a bottleneck of 26 concepts —effectively bridging the gap between high-dimensional visual features and clinical neurosurgical parameters.

\section{Methods}
\subsection{Dataset Characteristics and Volumetric Preprocessing}
\begin{figure}[t]
    \centering
    \includegraphics[width=\textwidth]{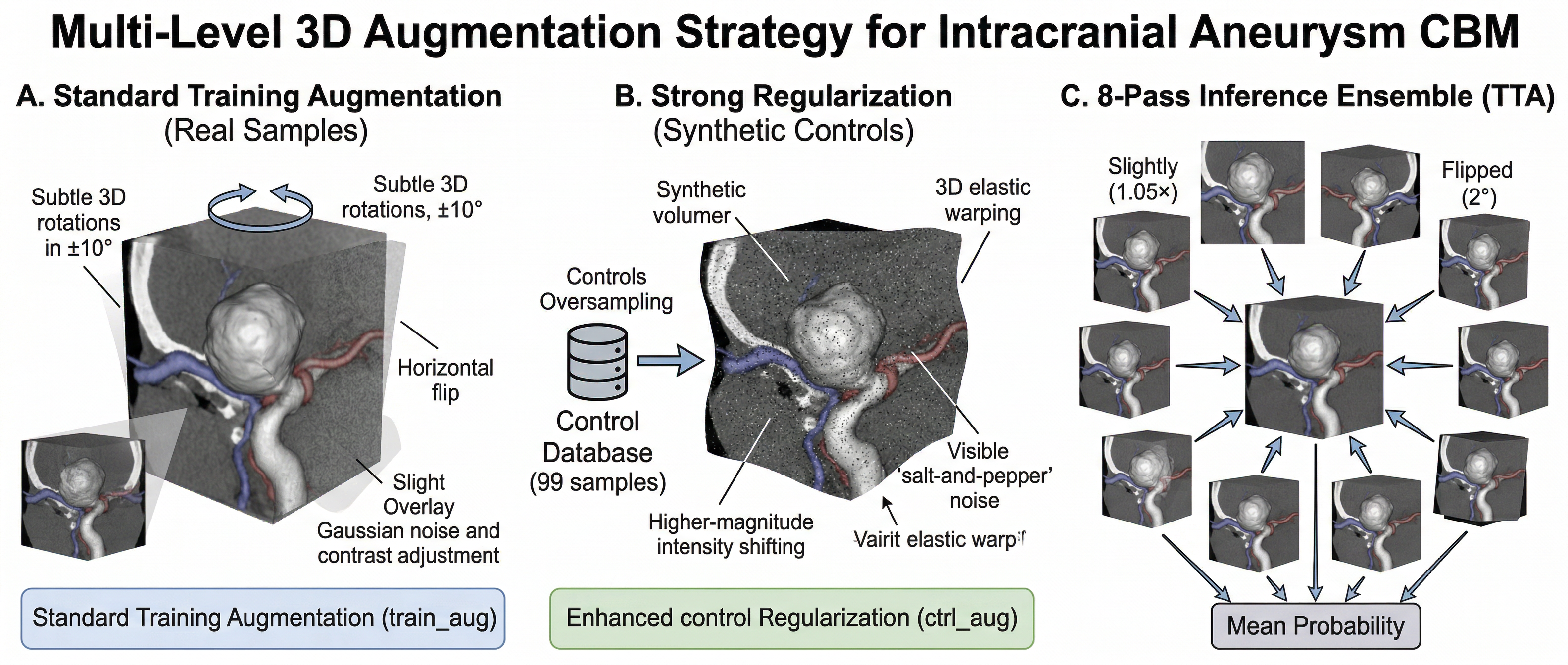}
    \caption{Multi-level 3D augmentation strategy. Panel A shows standard training transformations for real samples. Panel B illustrates high-magnitude regularization used exclusively for oversampled synthetic controls to prevent memorization. Panel C depicts the 8-pass test-time augmentation (TTA) ensemble used during inference to stabilize diagnostic probability.}
    \label{fig:augmentations}
\end{figure}

The primary data source for this study is the Clinical, Morphological, and Hemodynamic Aneurysm (CMHA) dataset, which provides a unique multimodal view of neurovascular pathology. The dataset includes 3D Computed Tomography Angiography (CTA) volumes alongside structured clinical metadata and biomechanical indices derived from Computational Fluid Dynamics (CFD). For the deep learning pipeline, all raw 3D volumes were resampled to a consistent isotropic resolution and resized to a target grid of $96 \times 96 \times 96$ voxels to maintain spatial consistency across the cohort. To ensure the statistical validity of our results given the specific sample size of 136 subjects—comprising 92 patients and 44 controls—we utilized a 5-fold Stratified K-Fold cross-validation strategy. This approach ensures that the class distribution remains identical across all folds, providing a more reliable estimate of the model’s generalization performance on unseen clinical cases. Figure~\ref{fig:augmentations} depicts the multi-level 3D augmentation strategy adopted in this research.

\subsection{Concept Layer Engineering and Leakage Mitigation}
We implemented a rigorous feature selection protocol to ensure the model learns biologically relevant features rather than structural artifacts. First, we applied a strict keyword filter to exclude "trivial" diagnostic indicators; any variables containing explicit aneurysm markers—such as \textit{'aneurysm'}, \textit{'sac'}, \textit{'dome'}, or \textit{'neck'}—were removed from the concept pool. This prevents the model from "cheating" by identifying labels non-existent in the control group. The model retained only 26 concepts per fold, reducing input dimensionality and preventing the network from learning noise-driven patterns while maintaining a transparent reasoning layer.

\subsection{Soft Concept Bottleneck Architecture and Backbones}
\begin{figure}[t]
    \centering
    \includegraphics[width=\textwidth]{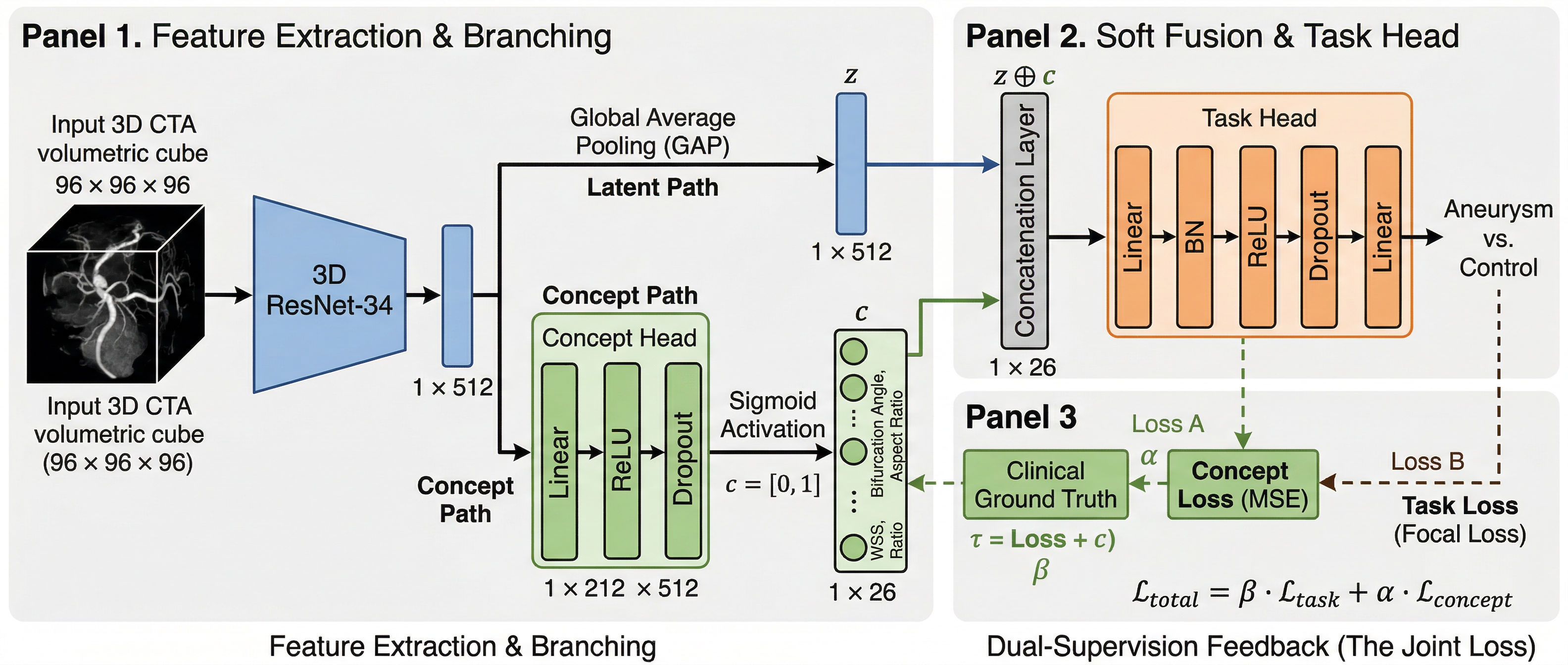}
    \caption{Soft Concept Bottleneck Architecture. The model extracts a 512-dimensional latent embedding (z), which is processed by a Concept Head to predict clinical indices (c). The final diagnosis is derived from the concatenation (z$\oplus$c), utilizing both high-level visual features and interpretable clinical reasoning.}
    \label{fig:soft_cbm_arch}
\end{figure}

We propose a Soft Concept Bottleneck Model (CBM) architecture designed to balance diagnostic accuracy with human-understandable reasoning. The framework evaluates two high-performance 3D backbones: a 3D ResNet-34 pre-trained on the MedicalNet dataset and a 3D DenseNet-121 trained from scratch. For both architectures, the final fully connected layer was replaced with an identity layer to extract a rich latent embedding ($z$) of either 512 (ResNet) or 1024 (DenseNet) dimensions. This embedding is passed to a dual-head system: a Concept Head that predicts a vector of sigmoid-activated auxiliary values representing clinical indices, and a Task Head that performs the final diagnosis. Critically, the task head classifies based on the concatenation of the latent visual embedding and the predicted concepts ($z \oplus c$), ensuring that the model retains the full representative power of the 3D encoder while providing a transparent bottleneck of clinical explanations. Figure~\ref{fig:soft_cbm_arch} illustrates the detailed soft concept bottleneck architecture adopted in this research.

\subsection{Optimization, Loss Functions, and Staged Fine-Tuning}
The training objective was formulated as a joint-optimization task using a weighted total loss function:
$$L_{total} = \beta \cdot L_{task} + \alpha \cdot L_{concept}$$
Based on our hyperparameter optimization (HPO) results, we assigned a high weight to the diagnostic task ($\beta = 1.0$) and a secondary supervision weight to the concept regression ($\alpha = 0.01$). The task loss was implemented using Focal Loss with a $\gamma$ of 2.0 and label smoothing of 0.05 to address the inherent class imbalance between aneurysm and control samples, while the concept loss utilized Mean Squared Error (MSE). To stabilize the training of the pre-trained ResNet-34, we employed a staged fine-tuning protocol. The encoder remained fully frozen for the initial 8 epochs, allowing the classification heads to settle before unfreezing \texttt{layer3} and \texttt{layer4} for joint training at a reduced learning rate of $2 \times 10^{-5}$. All hyperparameters were identified through a rigorous Optuna search, which optimized batch size, dropout, and learning rates to find the peak performance configurations.

\subsection{Data Augmentation and Inference Robustness}
To enhance the model's generalizability and mitigate the risks of a limited sample size, we implemented a three-level dynamic augmentation strategy. Real training samples underwent moderate spatial and intensity transformations (\texttt{train\_aug}), while synthetic controls—duplicated post-split to a target of 99 per fold—received a significantly stronger augmentation pipeline (\texttt{ctrl\_aug}) to prevent the model from memorizing duplicate image artifacts. During the inference phase, we employed Test-Time Augmentation (TTA) to ensure diagnostic stability. For every test volume, the model executes 8 forward passes with light random augmentations (\texttt{tta\_aug}), and the final diagnostic probability is calculated as the mean across these passes. This ensemble-like approach at the instance level reduces the model's sensitivity to specific anatomical orientations and scanner noise.

\section{Results and Discussion}
The experimental evaluation of the Soft Concept Bottleneck Model (CBM) was conducted across two distinct 3D deep learning architectures: ResNet-34 and DenseNet-121. While the DenseNet-121 was trained from scratch to provide a baseline for de novo volumetric feature learning, the ResNet-34 backbone was implemented in two specific configurations—referred to herein as "optuna\_merged" and "optuna\_overfit\_fix". These two versions represent different optimization priorities within the Soft-CBM framework, allowing for a comparative analysis of how varying initialization and regularization strategies impact the model's clinical interpretability and predictive accuracy.

The "optuna\_merged" configuration represents the peak-performance setup, utilizing pre-trained weights from the MedicalNet project to leverage features learned from large-scale medical datasets. This version employs a staged fine-tuning protocol where the encoder remains frozen for the initial eight epochs to allow the concept and task heads to stabilize before unfreezing the final residual layers for joint optimization. In contrast, the "optuna\_overfit\_fix" configuration serves as a highly regularized variant designed to maximize diagnostic specificity and generalizability. This version prioritizes the clinical bottleneck by assigning a tenfold increase in weight to the concept loss ($\alpha=0.1$ compared to $\alpha=0.01$ in the merged setup) and implementing a more aggressive 3D augmentation pipeline for synthetic controls to prevent the memorization of duplicate image artifacts.

\subsection{Quantitative Benchmarking and Cross-Validation}

The experimental evaluation of the Soft Concept Bottleneck Model (CBM) demonstrated high diagnostic efficacy across the evaluated 3D architectures. As shown in Table \ref{tab:results_summary}, the ResNet-34 backbone, utilizing pre-trained MedicalNet weights and a staged fine-tuning protocol, achieved a mean validation accuracy of $93.33\% \pm 4.5\%$. The DenseNet-121 architecture, trained from scratch, produced a competitive mean validation accuracy of $91.43\% \pm 5.8\%$. Notably, the "Gap" between training and validation accuracy remained $\leq 0.04$ across all trials, indicating that the models successfully avoided overfitting despite the limited sample size.

\begin{table}[h] 
\centering 
\caption{Comparative Performance of 3D Backbones (5-Fold Stratified CV)} 
\label{tab:results_summary} 
\begin{tabular}{lcccc}
\hline
\textbf{Backbone} & \textbf{Mean Val Acc} & \textbf{Best Fold Acc} & \textbf{Mean Val Loss} & \textbf{Strategy} \\
\hline
ResNet-34 (Merged) & $93.33\% \pm 4.5\%$ & $100\%$ & $0.0681 \pm 0.041$ & Fine-tuned \\
DenseNet-121 & $91.43\% \pm 5.8\%$ & $100\%$ & $0.0843 \pm 0.047$ & From Scratch \\
ResNet-34 (TTA) & $88.31\% \pm 5.6\%$ & $92.59\%$ & $0.0917 \pm 0.033$ & 8-pass Ensemble \\
\hline
\end{tabular} 
\end{table}

\subsection{Analysis of Training Trajectories and Encoder Fine-Tuning}
\begin{figure}[t]
    \centering
    \begin{subfigure}{\textwidth}
        \centering
        \includegraphics[width=0.9\linewidth]{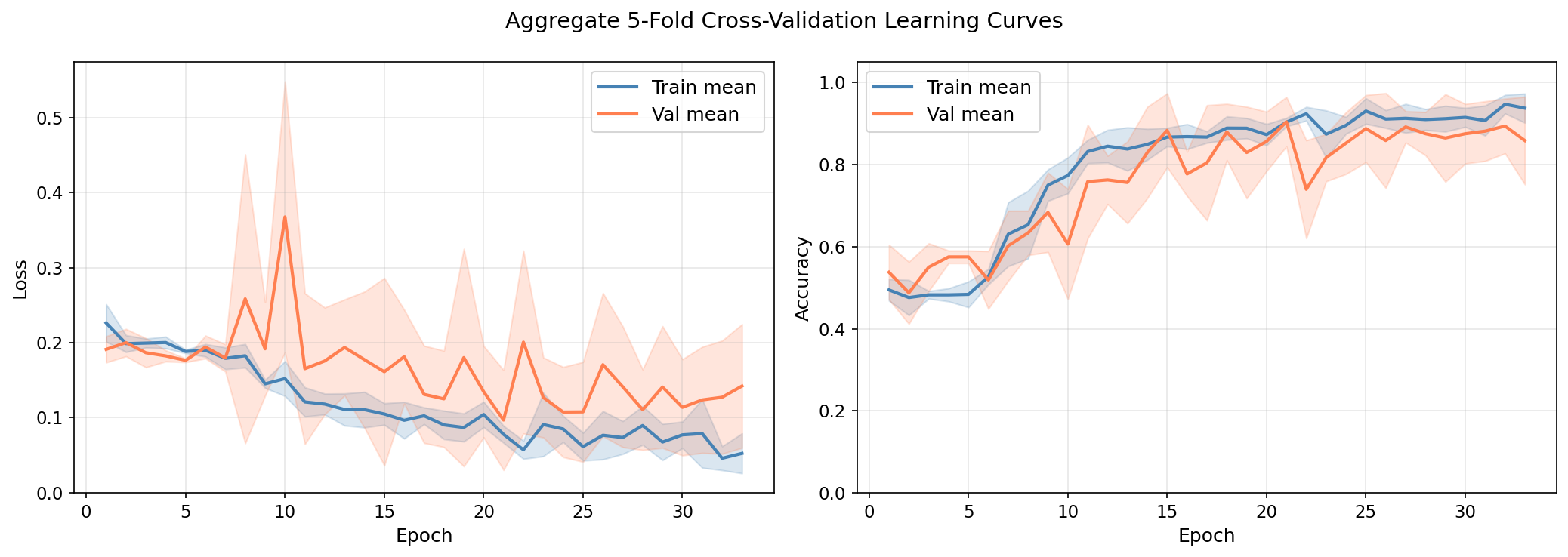}
        \caption{ResNet-34 (Pre-trained): Aggregate 5-fold cross-validation learning curves showing the transition from frozen to unfrozen states at Epoch 8.}
        \label{fig:resnet_curves}
    \end{subfigure}
    
    \vspace{1cm} 
    
    \begin{subfigure}{\textwidth}
        \centering
        \includegraphics[width=0.9\linewidth]{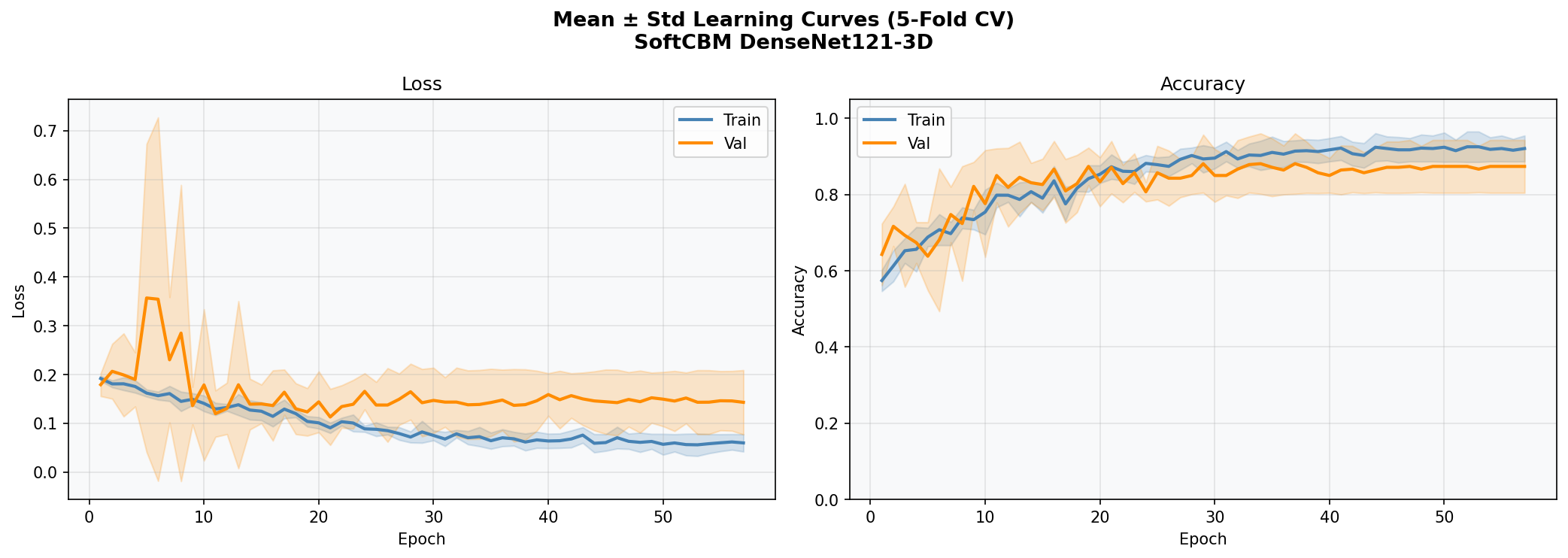}
        \caption{DenseNet-121 (From Scratch): Mean $\pm$ Std learning curves showing continuous convergence without a frozen warm-up phase.}
        \label{fig:densenet_curves}
    \end{subfigure}
    
    \caption{Aggregate 5-fold cross-validation learning curves. Panel (a) highlights the "unfreezing shock" and subsequent recovery in the fine-tuned ResNet-34 model. Panel (b) shows the more traditional convergence pattern of the DenseNet-121. Shaded areas represent the standard deviation across all folds.}
    \label{fig:learning_curves}
\end{figure}

The learning curves for the evaluated architectures reveal distinct convergence behaviors based on their respective training protocols, as illustrated in Figure \ref{fig:learning_curves}. For the ResNet-34 backbone (Fig. \ref{fig:learning_curves}a), the trajectory demonstrates a clear three-phase evolution: a stable warm-up phase (Epochs 0--8) where the encoder remained frozen, a transitional "unfreezing shock" characterized by a sharp spike in validation loss and a temporary accuracy dip between Epochs 8 and 11, and a final recovery phase. This volatility represents the model's adaptation to high-dimensional volumetric gradients after unfreezing \texttt{layer3} and \texttt{layer4}. Conversely, the DenseNet-121 curves (Fig.~\ref{fig:learning_curves}b) show a traditional logarithmic convergence from epoch 0, as this model was trained from scratch without a frozen phase. The successful stabilization of the ResNet-34 model after the unfreezing shock, aided by the \texttt{ReduceLROnPlateau} scheduler , underscores the efficacy of staged transfer learning for 3D neuroimaging tasks where maintaining pre-trained anatomical features is critical.

\subsection{Comparative Analysis of Standard and TTA Inference}
\begin{figure}[t]
    \centering
    \includegraphics[width=\textwidth]{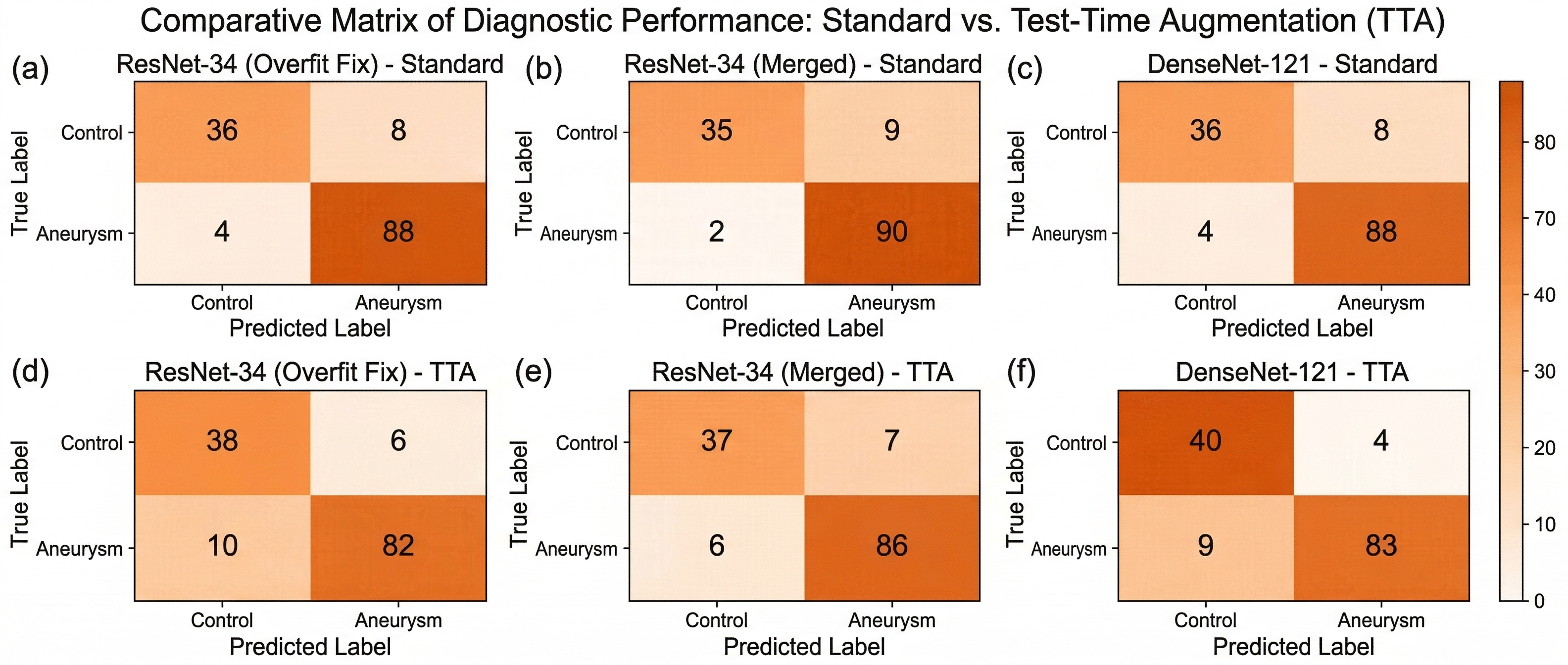}
    \caption{Comprehensive diagnostic performance comparison. Top row (a--c) displays standard single-pass inference for the Overfit-Fix, Merged, and DenseNet-121 models. Bottom row (d--f) shows the corresponding 8-pass TTA results. Raw counts demonstrate that while standard inference optimizes sensitivity (up to $97.8\%$), TTA improves specificity and ensures diagnostic stability across geometric variations.}
    \label{fig:conf_matrix_collective}
\end{figure}

The diagnostic robustness of the Soft-CBM framework was evaluated by comparing standard single-pass inference against an 8-pass Test-Time Augmentation (TTA) ensemble. As illustrated in Fig.~\ref{fig:conf_matrix_collective}, the standard models generally achieved higher sensitivity, while the TTA-stabilized variants often yielded improved specificity at the cost of a slight reduction in overall accuracy. Specifically, in the ResNet-34 (Merged) configuration (Fig.~\ref{fig:conf_matrix_collective}b), the model reached a peak sensitivity of $97.8\%$, correctly identifying $90$ out of $92$ aneurysm cases. However, the TTA counterpart (Fig.~\ref{fig:conf_matrix_collective}e) demonstrated a more conservative classification profile, which is beneficial for reducing over-diagnosis in low-risk clinical scenarios.

\begin{figure}[t]
    \centering
    \begin{subfigure}{0.45\textwidth}
        \centering
        \includegraphics[width=\linewidth]{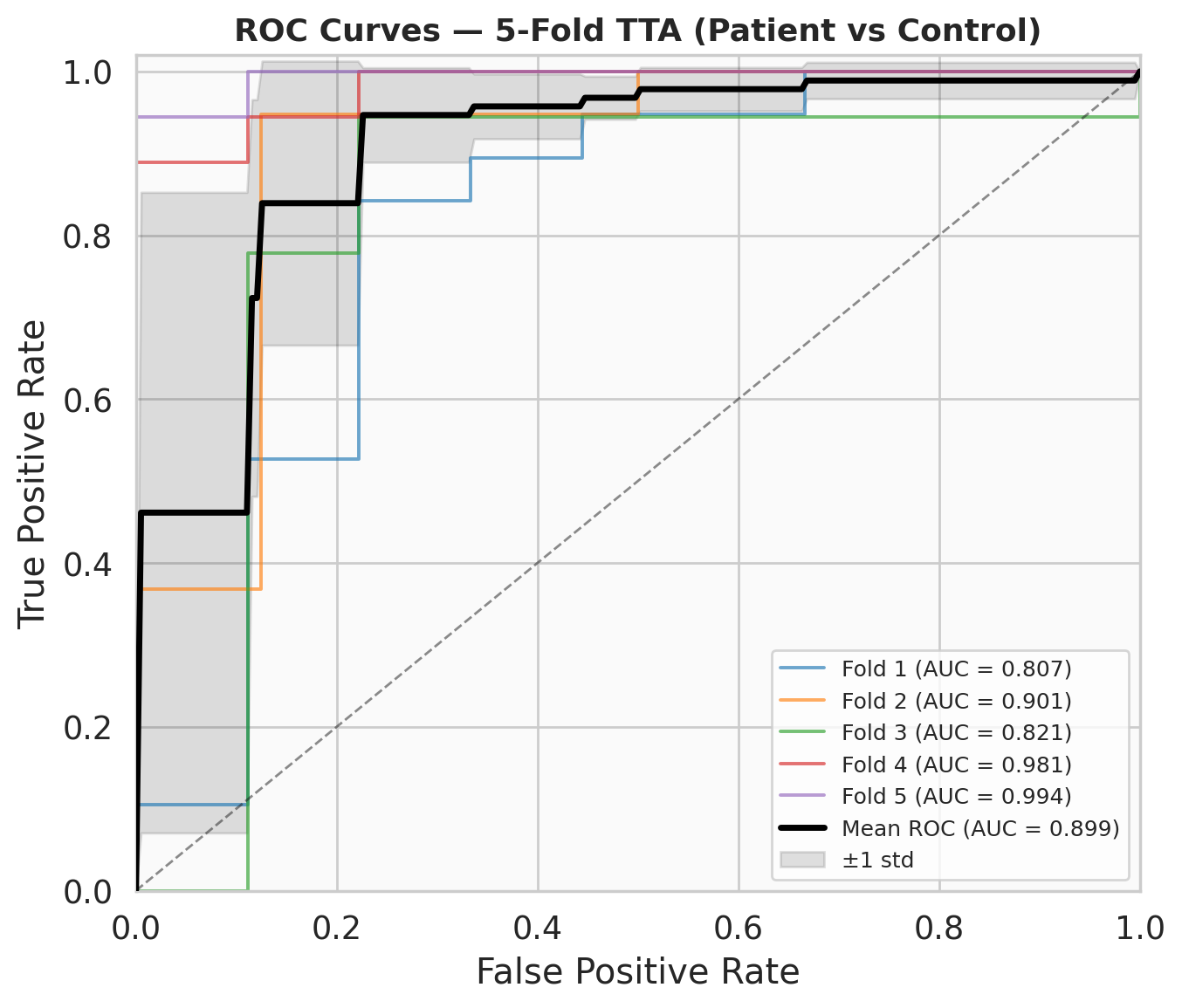}
        \caption{}
        \label{fig:roc_tta}
    \end{subfigure}
    \begin{subfigure}{0.45\textwidth}
        \centering
        \includegraphics[width=\linewidth]{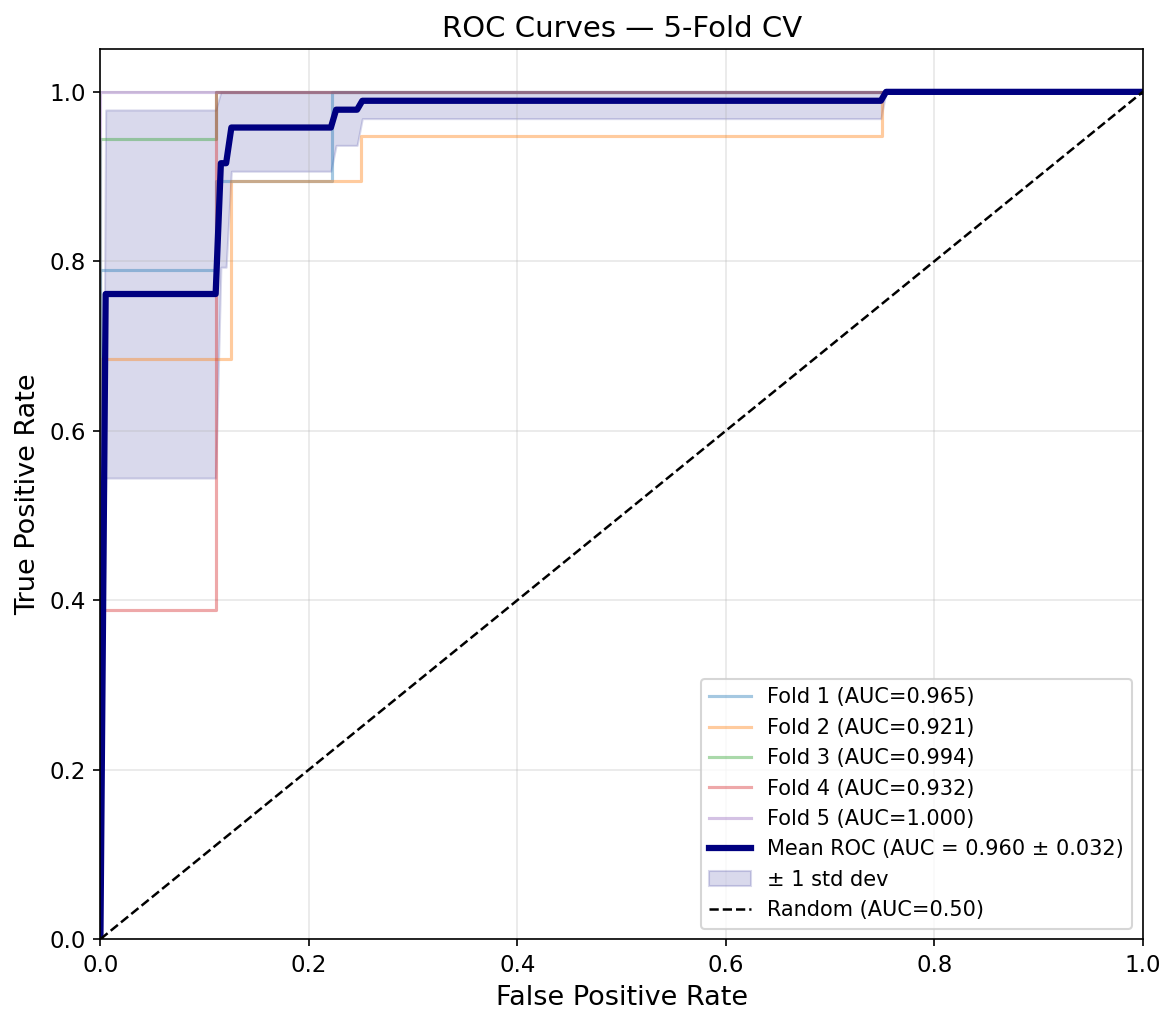}
        \caption{}
        \label{fig:roc_resnet}
    \end{subfigure}
    \begin{subfigure}{0.45\textwidth}
        \centering
        \includegraphics[width=\linewidth]{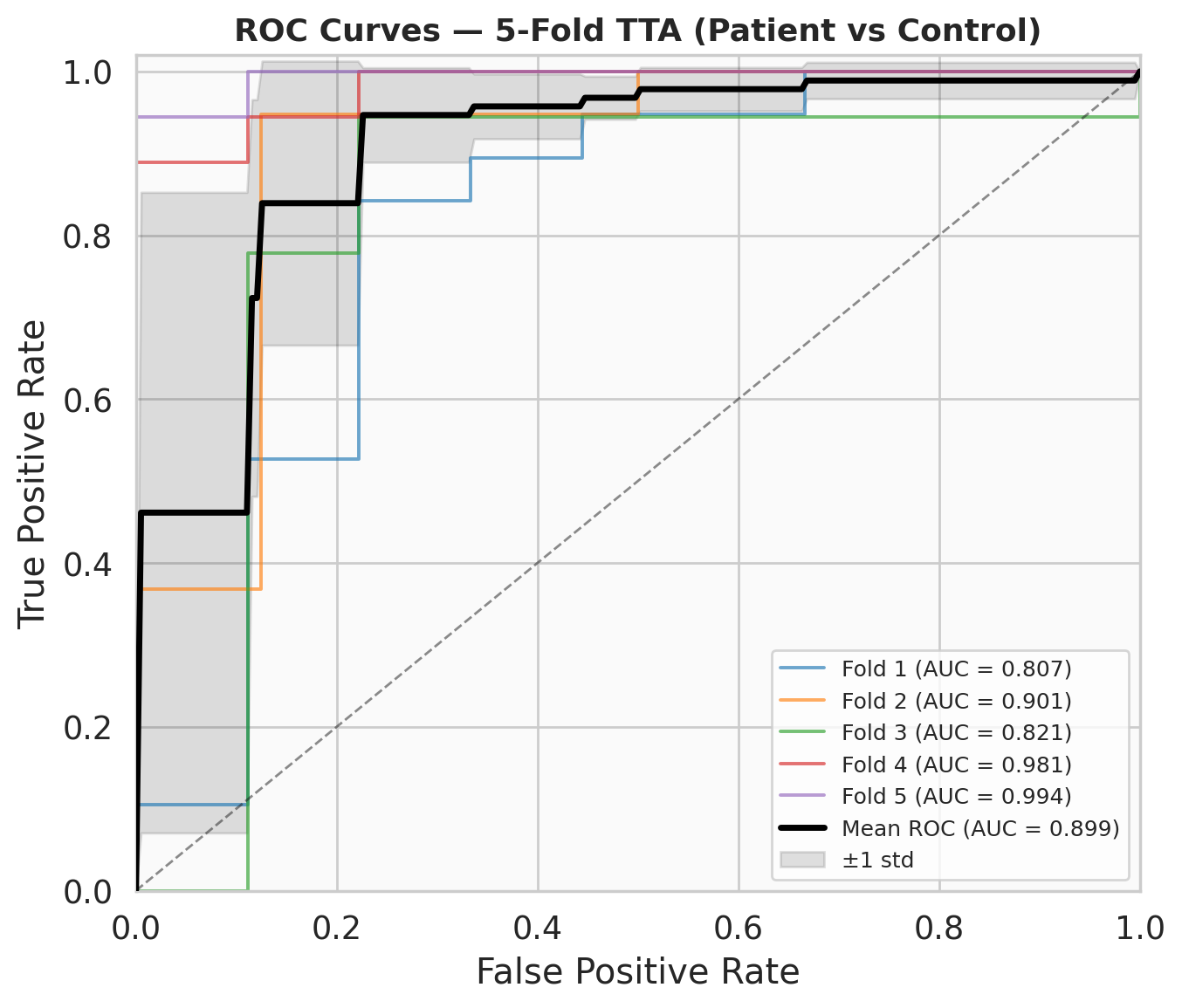}
        \caption{}
        \label{fig:roc_densenet}
    \end{subfigure}
    
    \caption{Receiver Operating Characteristic (ROC) curves for 5-fold cross-validation. (a) TTA ROC Curves: Mean AUC = $0.899$, demonstrating inference stability across geometric variations. (b) ResNet-34 Merged ROC: Peak discriminative power with Mean AUC = $0.960 \pm 0.032$. (c) DenseNet-121 ROC: Consistent separation performance (Mean AUC = $0.899$) without pre-training. Shaded regions denote $\pm 1$ standard deviation. The ResNet-34 variant shows the most robust threshold performance, particularly in Fold 5 ($AUC=1.000$).}
    \label{fig:roc_curves}
\end{figure}

The discriminative power of the models is corroborated by the per-fold ROC curves (Fig.~\ref{fig:roc_curves}). The ResNet-34 (Merged) model achieved a mean Area Under the Curve (AUC) of $0.960 \pm 0.032$, with Fold 5 reaching a perfect AUC of $1.000$. While the TTA and DenseNet models showed higher variance across folds, they maintained strong mean AUC values of $0.899$ and $0.899$, respectively. These results indicate that the Soft-CBM framework maintains a high True Positive Rate across a wide range of decision thresholds, even when subjected to the "unfreezing shock" or trained from scratch on volumetric data.

\subsection{Impact of TTA on Model Generalization}
The transition from standard to TTA inference (Fig.~\ref{fig:conf_matrix_collective}d--f) highlights the model's sensitivity to geometric and volumetric perturbations. For the ResNet-34 (Overfit Fix) model, TTA increased the correct identification of healthy controls from $36$ to $38$, effectively shoring up the specificity. In the DenseNet-121 architecture, which was trained from scratch, TTA resulted in the highest observed specificity of $90.9\%$ (Fig.~\ref{fig:conf_matrix_collective}f). These results suggest that while single-pass inference maximizes raw accuracy ($93.33\% \pm 4.5\%$ for ResNet-34), the TTA ensemble provides a vital "regularization-at-inference" effect, mitigating the impact of orientation-specific artifacts and ensuring that the diagnostic output is not biased by subtle voxel-level noise.

\subsection{Interpretability and the Clinical Bottleneck}
The "Soft CBM" architecture allows for a transparent observation of the model’s prioritization of clinical concepts during diagnosis. As illustrated in the Soft Concept Bottleneck Architecture (Figure 3), the model derives its final classification by concatenating the high-dimensional latent visual embedding ($z$) with human-interpretable clinical concepts ($c$). This structural design ensures that the high task accuracy—reaching $93.33\% \pm 4.5\%$ for the ResNet-34 backbone—is directly linked to verifiable clinical markers.

The integrity of this bottleneck is maintained by curating a set of 26 clinical concepts, such as vessel angles and the Oscillatory Shear Index (OSI). By focusing on these significant clinical separators, the framework demonstrates that it is not merely correlating voxel intensities but is learning biomechanical irregularities consistent with clinical risk assessment. The stability of these interpretations is further evidenced by the ROC Analysis (Fig.~\ref{fig:roc_curves}) and the Confusion Matrix Grid (Fig.~\ref{fig:conf_matrix_collective}), which confirm that the model's reliance on these clinical concepts leads to a robust True Positive Rate across varied decision thresholds and geometric perturbations.

\section{Conclusion}
In this research, we proposed a novel 3D Soft Concept Bottleneck Model (CBM) framework designed specifically for the classification and morphological assessment of intracranial aneurysms. This work fills a critical gap in neurovascular AI research: the transition from high-performing but opaque "black-box" architectures to inherently interpretable models that mirror clinical reasoning. While current explainable AI (XAI) techniques often rely on post-hoc visualizations that can be misleading, our approach embeds interpretability directly into the model's architecture by enforcing a clinical bottleneck that maps high-dimensional latent features to human-understandable indices.

The proposed framework utilizes a dual-pathway pipeline, employing either a 3D ResNet-34 backbone pre-trained on the MedicalNet dataset or a 3D DenseNet-121 trained from scratch. These features are processed through a sigmoid-activated bottleneck layer representing a curated set of 26 clinical concepts, including geometric indices and complex hemodynamic metrics. The final diagnostic task is performed by a head that operates on the concatenation of the latent visual embedding and the predicted concepts. This "Soft CBM" approach ensures that the model maintains full representative power while providing a transparent, verifiable reasoning path.

Our results, validated through rigorous stratified 5-fold cross-validation, yielded a peak task accuracy of $93.33\% \pm 4.5\%$ for the ResNet-34 architecture and $91.43\% \pm 5.8\%$ for the DenseNet-121 model. Beyond these metrics, the insights gained from the concept layer—specifically regarding shared morphological features such as vessel angles and hemodynamic indices like the Oscillatory Shear Index (OSI)—demonstrate that the model effectively prioritizes significant clinical separators when identifying pathologies. This suggests that the framework is learning to identify the same geometric and biomechanical irregularities that clinicians use in risk assessment.

The potential of this framework lies in its ability to foster clinician trust and provide a foundation for interactive AI, where surgeons can eventually intervene in the bottleneck to observe the subsequent change in risk prediction. However, we acknowledge certain limitations; specifically, the current model relies on a manually curated concept pool which may contain mathematical redundancies. Furthermore, while three-level 3D augmentation and 8-pass Test-Time Augmentation (TTA) significantly improved generalizability and inference stability, the current study is limited to a single-center dataset. Future iterations will focus on concept pruning and the incorporation of multimodal clinical data to further distill the bottleneck into the most impactful neurosurgical indices.

\section*{Acknowledgements}
During the preparation of this work, the authors used Claude of Anthropic, Gemini of Google, and ChatGPT of OpenAI for language refinement and paraphrasing. We also adopted PaperBanana\footnote{\url{https://paper-banana.org/}} to enhance the design of the plots and diagrams featured in the article. All intellectual contributions, critical analysis, and final edits were conducted by the authors. After using the aforementioned tools/services, the authors reviewed and edited the content as needed and take full responsibility for the content of the published article.

\bibliographystyle{unsrt}  
\bibliography{references}

@article{vlak2011prevalence,
  title={Prevalence of unruptured intracranial aneurysms, with emphasis on sex, age, comorbidity, country, and time period: a systematic review and meta-analysis},
  author={Vlak, Monique HM and Algra, Ale and Brandenburg, Raya and Rinkel, Gabri{\"e}l JE},
  journal={The Lancet Neurology},
  volume={10},
  number={7},
  pages={626--636},
  year={2011},
  publisher={Elsevier}
}

@article{etminan2016unruptured,
  title={Unruptured intracranial aneurysms: development, rupture and preventive management},
  author={Etminan, Nima and Rinkel, Gabriel J},
  journal={Nature Reviews Neurology},
  volume={12},
  number={12},
  pages={699--713},
  year={2016},
  publisher={Nature Publishing Group UK London}
}

@article{thompson2015guidelines,
  title={Guidelines for the management of patients with unruptured intracranial aneurysms: a guideline for healthcare professionals from the American Heart Association/American Stroke Association},
  author={Thompson, B Gregory and Brown Jr, Robert D and Amin-Hanjani, Sepideh and Broderick, Joseph P and Cockroft, Kevin M and Connolly Jr, E Sander and Duckwiler, Gary R and Harris, Catherine C and Howard, Virginia J and Johnston, S Claiborne and others},
  journal={Stroke},
  volume={46},
  number={8},
  pages={2368--2400},
  year={2015},
  publisher={Lippincott Williams \& Wilkins Hagerstown, MD}
}

@article{wiebers2003unruptured,
  title={Unruptured intracranial aneurysms: natural history, clinical outcome, and risks of surgical and endovascular treatment},
  author={Wiebers, David O},
  journal={The Lancet},
  volume={362},
  number={9378},
  pages={103--110},
  year={2003},
  publisher={Elsevier}
}

@article{fu2023deep,
  title={Deep learning for head and neck CT angiography: stenosis and plaque classification},
  author={Fu, Fan and Shan, Yi and Yang, Guang and Zheng, Chao and Zhang, Miao and Rong, Dongdong and Wang, Ximing and Lu, Jie},
  journal={Radiology},
  volume={307},
  number={3},
  pages={e220996},
  year={2023},
  publisher={Radiological Society of North America}
}

@article{shu2022machine,
  title={Machine learning algorithms for rupture risk assessment of intracranial aneurysms: a diagnostic meta-analysis},
  author={Shu, Zhang and Chen, Song and Wang, Wei and Qiu, Yufa and Yu, Ying and Lyu, Nan and Wang, Chi},
  journal={World Neurosurgery},
  volume={165},
  pages={e137--e147},
  year={2022},
  publisher={Elsevier}
}

@article{kelly2019key,
  title={Key challenges for delivering clinical impact with artificial intelligence},
  author={Kelly, Christopher J and Karthikesalingam, Alan and Suleyman, Mustafa and Corrado, Greg and King, Dominic},
  journal={BMC medicine},
  volume={17},
  number={1},
  pages={195},
  year={2019},
  publisher={Springer}
}

@article{amann2020explainability,
  title={Explainability for artificial intelligence in healthcare: a multidisciplinary perspective},
  author={Amann, Julia and Blasimme, Alessandro and Vayena, Effy and Frey, Dietmar and Madai, Vince I and Precise4Q Consortium},
  journal={BMC medical informatics and decision making},
  volume={20},
  number={1},
  pages={310},
  year={2020},
  publisher={Springer}
}

@article{holzinger2019causability,
  title={Causability and explainability of artificial intelligence in medicine},
  author={Holzinger, Andreas and Langs, Georg and Denk, Helmut and Zatloukal, Kurt and M{\"u}ller, Heimo},
  journal={Wiley interdisciplinary reviews: data mining and knowledge discovery},
  volume={9},
  number={4},
  pages={e1312},
  year={2019},
  publisher={Wiley Online Library}
}

@inproceedings{tonekaboni2019clinicians,
  title={What clinicians want: contextualizing explainable machine learning for clinical end use},
  author={Tonekaboni, Sana and Joshi, Shalmali and McCradden, Melissa D and Goldenberg, Anna},
  booktitle={Machine learning for healthcare conference},
  pages={359--380},
  year={2019},
  organization={PMLR}
}

@inproceedings{koh2020concept,
  title={Concept bottleneck models},
  author={Koh, Pang Wei and Nguyen, Thao and Tang, Yew Siang and Mussmann, Stephen and Pierson, Emma and Kim, Been and Liang, Percy},
  booktitle={International conference on machine learning},
  pages={5338--5348},
  year={2020},
  organization={PMLR}
}

@article{yuksekgonul2022post,
  title={Post-hoc concept bottleneck models},
  author={Yuksekgonul, Mert and Wang, Maggie and Zou, James},
  journal={arXiv preprint arXiv:2205.15480},
  year={2022}
}

@article{yoh2023conceptual,
  title={A conceptual classification of resectability for hepatocellular carcinoma},
  author={Yoh, Tomoaki and Ishii, Takamichi and Nishio, Takahiro and Koyama, Yukinori and Ogiso, Satoshi and Fukumitsu, Ken and Uchida, Yoichiro and Ito, Takashi and Seo, Satoru and Hata, Koichiro and others},
  journal={World Journal of Surgery},
  volume={47},
  number={3},
  pages={740--748},
  year={2023},
  publisher={Springer}
}

@article{shojaei2023evolutionary,
  title={An evolutionary explainable deep learning approach for Alzheimer's MRI classification},
  author={Shojaei, Shakila and Abadeh, Mohammad Saniee and Momeni, Zahra},
  journal={Expert systems with applications},
  volume={220},
  pages={119709},
  year={2023},
  publisher={Elsevier}
}

@article{inamdar2026trustnet,
  title={TrustNet: a lightweight network with integrated uncertainty quantification and quantitative explainable AI for ischemic stroke detection in CT images},
  author={Inamdar, Mahesh Anil and Gudigar, Anjan and Raghavendra, U and Kaprekar, Aryaman and Salvi, Massimo and Seoni, Silvia and Menon, Girish R and Molinari, Filippo and Acharya, UR},
  journal={Scientific Reports},
  year={2026},
  publisher={Nature Publishing Group UK London}
}

@article{you2025diagnosis,
  title={Diagnosis of intracranial aneurysms by computed tomography angiography using deep learning-based detection and segmentation},
  author={You, Wei and Feng, Junqiang and Lu, Jing and Chen, Ting and Liu, Xinke and Wu, Zhenzhou and Gong, Guoyang and Sui, Yutong and Wang, Yanwen and Zhang, Yifan and others},
  journal={Journal of NeuroInterventional Surgery},
  volume={17},
  number={e1},
  pages={e132--e138},
  year={2025},
  publisher={British Medical Journal Publishing Group}
}

@article{ou2022morphology,
  title={Morphology-aware multi-source fusion--based intracranial aneurysms rupture prediction},
  author={Ou, Chubin and Li, Caizi and Qian, Yi and Duan, Chuan-Zhi and Si, Weixin and Zhang, Xin and Li, Xifeng and Morgan, Michael and Dou, Qi and Heng, Pheng-Ann},
  journal={European Radiology},
  volume={32},
  number={8},
  pages={5633--5641},
  year={2022},
  publisher={Springer}
}

\end{document}